\documentclass[conference,letterpaper]{IEEEtran}
\usepackage[T1]{fontenc}
\usepackage[utf8]{inputenc}
\usepackage{newtxtext}
\usepackage{newtxmath}

\usepackage{amsmath,amssymb,amsfonts,mathtools}
\usepackage{bm}
\usepackage{booktabs,multirow}
\usepackage{tabularx}
\usepackage{graphicx}
\usepackage{svg}
\graphicspath{{./Figures/}{./figs/}}
\usepackage[caption=false,font=footnotesize]{subfig}

\usepackage{enumitem}
\usepackage{xcolor}
\usepackage{siunitx}
\usepackage{algorithm}
\usepackage{algorithmic}
\usepackage{cite}
\usepackage[hidelinks]{hyperref}
\usepackage[nameinlink]{cleveref}

\newif\ifshowrevnotes

\showrevnotesfalse
\def\stripheader#1: #2\endmarker{#2}
\newif\ifshowcitetitles
\showcitetitlesfalse
\newif\ifjournal
\journalfalse
\newif\ifanonymous
\anonymousfalse 
\newcommand{\XhR}{\widehat{\mathrm{RW}}}

\newcommand{\eps}{\varepsilon}

\newcommand{\R}{\mathbb{R}}
\newcommand{\Gth}{G_{\Theta}}
\newcommand{\RW}[1]{\ensuremath{\mathrm{RW}\!\bigl(#1\bigr)}}
\newcommand{\Loss}{\mathcal{L}}

\DeclareMathOperator*{\argmin}{arg\,min}

\DeclareMathOperator{\Dist}{D}
\DeclareMathOperator{\Din}{D_{\mathrm{in}}}
\DeclareMathOperator{\Dout}{D_{\mathrm{out}}}
\DeclareMathOperator{\Halg}{H} 

\newcommand{\repourl}{https://github.com/Danial-Safaei/DFF}
\newcommand{\repo}{\ifanonymous\textit{(link hidden for review)}\else\href{\repourl}{\texttt{DFF-repo}}\fi}
\newcommand{\epsSteer}{\num[scientific-notation=false]{0.087}}
\newcommand{\epsDA}{\num[scientific-notation=false]{0.699}}
\newcommand{\epsLL}{\num[scientific-notation=false]{0.736}}
\newcommand{\epsSteerNinety}{\num[scientific-notation=false]{0.064}}
\newcommand{\epsDANinety}{\num[scientific-notation=false]{0.570}}
\newcommand{\epsLLNinety}{\num[scientific-notation=false]{0.582}}

\title{Quantifying Fidelity: A Decisive Feature Approach to Comparing Synthetic and Real Imagery}

\IEEEspecialpapernotice{\textcolor{red}{%
Preprint version of IEEE IV'26. Accepted by IV 2026: \href{https://ieee-iv.org/2026/}{https://ieee-iv.org/2026/}.%
}}

\author{\IEEEauthorblockN{Danial Safaei\IEEEauthorrefmark{1},
Siddartha Khastgir\IEEEauthorrefmark{1},
Mohsen Alirezaei\IEEEauthorrefmark{2},
Jeroen Ploeg\IEEEauthorrefmark{2},
Chih-Hong Cheng\IEEEauthorrefmark{3}, \\
Son Tong\IEEEauthorrefmark{2} and
Xingyu Zhao\IEEEauthorrefmark{1}}
\IEEEauthorblockA{\IEEEauthorrefmark{1}WMG, University of Warwick, Coventry, U.K.\\
\{Danial.Safaei, S.Khastgir.1, Xingyu.Zhao\}@warwick.ac.uk}
\IEEEauthorblockA{\IEEEauthorrefmark{2}Siemens Digital Industries Software\\
\{mohsen.alirezaei, son.tong\}@siemens.com, jeroen@2getthere.eu}
\IEEEauthorblockA{\IEEEauthorrefmark{3}Carl von Ossietzky Universität Oldenburg, Oldenburg, Germany\\
chih-hong.cheng@uni-oldenburg.de}
\thispagestyle{empty}
}

\begin{document}
\maketitle
\thispagestyle{empty}
\pagestyle{empty}

\begin{abstract}
Virtual testing using synthetic data has become a cornerstone of autonomous vehicle (AV) safety assurance. Despite progress in improving visual realism through advanced simulators and generative AI, recent studies reveal that pixel-level fidelity alone does not ensure reliable transfer from simulation to the real world. What truly matters is whether the system-under-test (SUT) bases its decisions on consistent \emph{decision evidence} in both real and simulated environments, not just whether images ``look real'' to humans.
To this end this paper proposes a behavior-grounded fidelity measure by introducing Decisive Feature Fidelity (DFF), a new SUT-specific metric that extends the existing fidelity spectrum to capture \emph{mechanism parity}, that is, agreement in the \emph{model-specific decisive evidence} that drives the SUT's decisions across domains.
DFF leverages explainable-AI methods to identify and compare the decisive features driving the SUT’s outputs for matched real–synthetic pairs. We further propose estimators based on counterfactual explanations, along with a DFF-guided calibration scheme to enhance simulator fidelity.
Experiments on 2126 matched KITTI–VirtualKITTI2 pairs demonstrate that DFF reveals discrepancies overlooked by conventional output-value fidelity. Furthermore, results show that DFF-guided calibration improves decisive-feature and input-level fidelity without sacrificing output value fidelity across diverse SUTs.
\end{abstract}
\section{Introduction}
\label{sec:intro}

Scenario-based virtual testing with synthetic data is a standard for AV safety evaluation \cite{iso_scenario_2022,bsi_scenario_2023}, enabling the recreation of complex driving situations.
Substantial resources are being directed toward high-fidelity simulators \cite{dosovitskiyCARLAOpenUrban2017} and generative-AI methods \cite{rombachHighResolutionImageSynthesis2022,zhangAddingConditionalControl2023} to improve realism. Most existing efforts focus on ``pixel-level realism'', aligning synthetic images with real images in a way that is agnostic to the system-under-test (SUT).
However, recent studies indicate that pixel-level similarity does not ensure reliable transfer of assessment conclusions between virtual and real domains \cite{chengInstanceLevelSafetyAwareFidelity2024,stoccoMindGapStudy2023,johnsonLiteratureReviewSimulation2023,zhao_statistical_2025}. What matters is whether the SUT bases its decisions on consistent visual evidence across domains.
This has shifted attention from appearance-based fidelity toward behavior-grounded validity and SUT-specific fidelity notions \cite{Riedmaier2020Survey,Riedmaier2022Taxonomy,Tang2023ADSTesting,Omeiza2022XAI}. In this paper, we focus on image inputs for brevity, but the framework generalizes to other modalities by replacing pixel correspondence with modality-appropriate correspondence.

\begin{figure*}[!t]
  \centering
  \includegraphics[width=2\columnwidth]{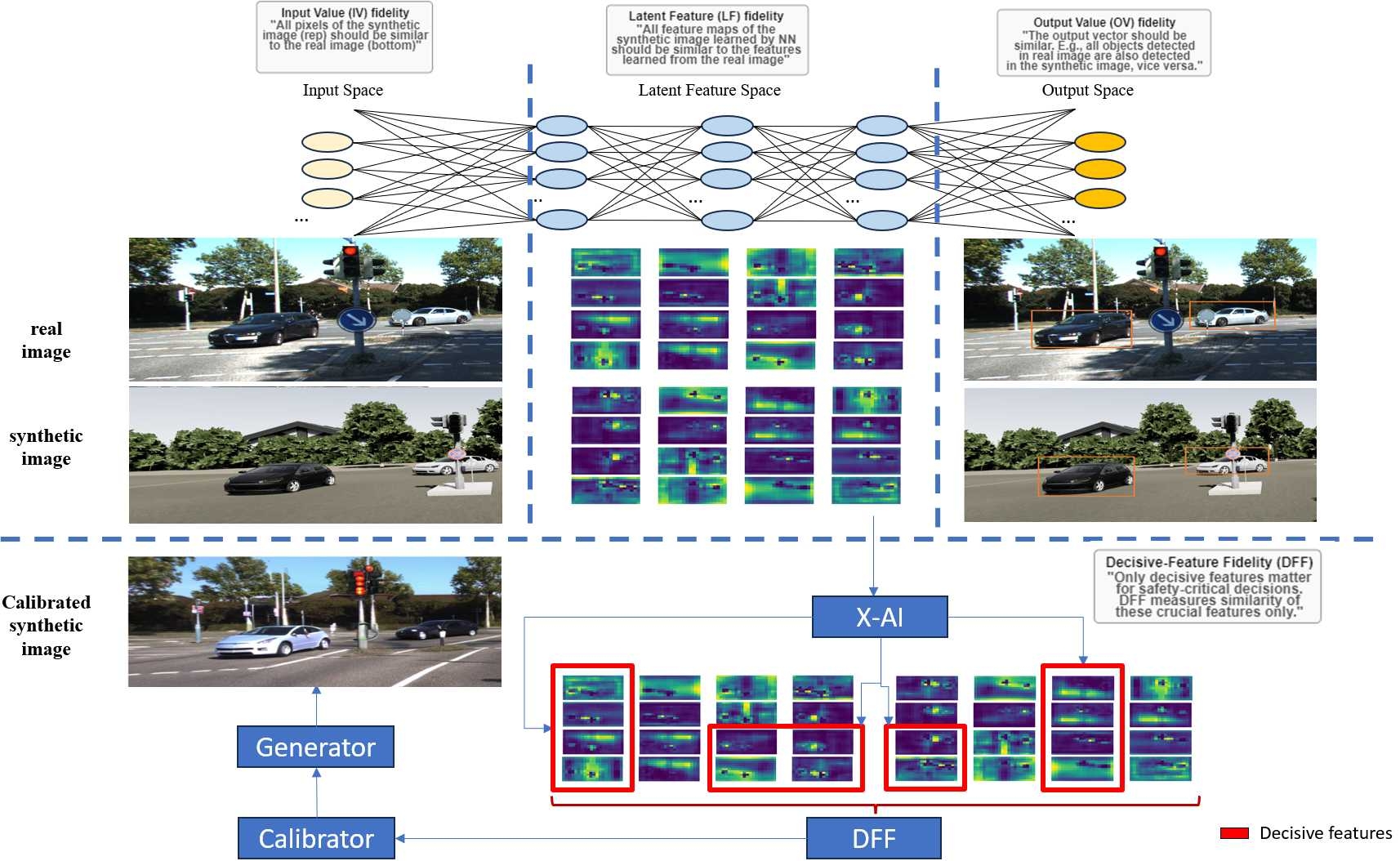}
  \caption{Top: three levels of the fidelity spectrum (IV, LF, OV) from~\cite{chengInstanceLevelSafetyAwareFidelity2024}. Bottom: our extension that adds decisive-feature fidelity (DFF) and a calibrator to adjust the generator. Safety-aware fidelity (SA) is defined in~\cite{chengInstanceLevelSafetyAwareFidelity2024} but is not explicitly visualized here.}
  \label{fig:dff-pipeline}
\end{figure*}

Previous work~\cite{chengInstanceLevelSafetyAwareFidelity2024} systematically categorized synthetic-data fidelity into four levels, as shown in Fig.~\ref{fig:dff-pipeline}: input-level (IV) fidelity, latent-feature (LF) fidelity, output-value (OV) fidelity, and safety-aware (SA) fidelity. As the names suggest, these correspond respectively to the fidelity measured in the SUT’s input space (e.g., the perception neural network’s raw inputs), latent feature space, output vector space, and safety-critical output space. While this spectrum of fidelity spans the entire processing pipeline of a neural network, it is too coarse to capture \textit{why} the SUT exhibits the same or different behaviors when presented with paired real and synthetic inputs representing the same scenario. For example, an AV perception module may output the same steering command, say a 20-degree left turn, for both the real and synthetic inputs, implying high OV-fidelity. Yet \textit{the underlying reasons could differ}: the real image might trigger the action due to a detected pedestrian ahead, whereas in the synthetic version of the same scenario, the same decision arises from detecting a right-turn sign next to the dog (and the dog is misdetected, which is concerning). In this case, the system produces the same decision for different reasons, i.e., based on different causal evidence. Notably, the LF-fidelity defined in~\cite{chengInstanceLevelSafetyAwareFidelity2024} cannot fully capture such discrepancies, as it simply compares all extracted features without distinguishing those that actually influence the SUT’s decision. This reveals a gap: existing metrics tell us \emph{what} the SUT outputs, but not \emph{why}. We need a metric that checks whether the SUT uses the same visual evidence (e.g., road edges, lane markings) in both domains. This gap directly motivates DFF.
The top part of Fig.~\ref{fig:dff-pipeline} recalls the input/latent/output/safety-aware fidelity spectrum of \cite{chengInstanceLevelSafetyAwareFidelity2024}. The bottom part illustrates our extension: an  Explainable AI (XAI) explainer extracts decisive features from the SUT for each paired real and synthetic input, the distance between decisive maps yields a Decisive Feature Fidelity (DFF) estimate, and this estimate is feedback to a calibrator that adjusts generator parameters to close mechanism gaps.
Throughout this paper, we evaluate and calibrate synthetic data for \emph{testing} a fixed SUT; we do not train or adapt the SUT itself.
To bridge this gap, we propose an explainable-AI (XAI)-based fidelity metric that \textit{quantifies the differences in the decisive features influencing the SUT’s outputs} for a given pair of real and synthetic data. Specifically, we first formalize a new SUT-specific fidelity metric, termed Decisive Feature Fidelity (DFF), which extends the fidelity spectrum defined in \cite{chengInstanceLevelSafetyAwareFidelity2024}.
Unlike the LF-fidelity in \cite{chengInstanceLevelSafetyAwareFidelity2024}, which compares all extracted features, DFF uses recent XAI methods to estimate and compare only the (subset of) features that truly influence the SUT’s output.
We then design practical estimators based on counterfactual explanations to quantify DFF and use these estimates as feedback signals to \textit{recalibrate} the synthetic data generator, thereby improving fidelity.
% In our experiments we instantiate $G$ with a diffusion-based generator, but the proposed calibration objective is general and applies equally to reconfigurable simulators such as CARLA or PreScan.
In our experiments we use a diffusion-based image generator, but the proposed calibration objective is general and also applies to reconfigurable simulators such as CARLA or Prescan.
To validate our apphroach, we conduct experiments on 2126 matched KITTI--VirtualKITTI2 pairs across three SUTs (steering, YOLOP drivable-area, and YOLOP lane-line), showing the utility of DFF for both assessment and calibration. DFF is most appropriate in \emph{clone/regression} regimes where the simulator aims to reproduce a real scene (or a nahrrowly defined scenario description) and the goal is to ensure that synthetic test conclusions transfer to the corresponding real setting.
In \emph{counterfactual stress testing} regimes, simulation is intentionally modified (e.g., actors moved, hazards injected, weather changed), and strict alignment to a particular real frame is not necessarily desirable.
In such cases, DFF is better used as (i) a \emph{within-regime} consistency check (mechanism stability across controlled perturbations), or (ii) a \emph{distributional} constraint to keep decisive evidence within an acceptable envelope rather than matching a specific real instance.

In summary, the main contributions include:
\begin{itemize}[leftmargin=*,noitemsep,topsep=0pt]
  \item \textbf{A new SUT-specific fidelity metric}: We formalize DFF, a metric that measures mechanism parity. We contrast it with existing metrics, arguing that focusing on the \emph{decisive-feature subspace} is more meaningful than comparing raw inputs (IV), entire latent spaces (LF), or just final outputs (OV) of the SUT.
  \item \textbf{An XAI-based DFF estimator}: We develop a practical estimator, using XAI methods such as counterfactual explainers (CF-XAI). The decisive maps for real and synthetic inputs are computed independently and then compared within this subspace.
  \item \textbf{A DFF-guided calibration objective}: We formulate an optimization objective that utilizes DFF estimates as feedback signals to fine-tune the synthetic data generator. Validation on a held-out split demonstrates that this approach improves decisive-feature and input-level fidelity. Crucially, it preserves behavioral (OV) fidelity, ensuring that the SUT's task performance does not degrade beyond a strictly defined tolerance (a pre-declared non-inferiority margin).
\end{itemize}

\section{Related Work}
\label{sec:related}
Virtual testing offers a scalable and risk-free way to assess AV safety across diverse scenarios. However, the fidelity of synthetic data used in such testing remains a critical and actively studied issue due to the persistent sim-to-real gap~\cite{10242366,10588858,dosovitskiyCARLAOpenUrban2017,geigerAreWeReady2012}. While fidelity can be defined and classified in multiple ways, an important distinction lies between SUT-agnostic and SUT-specific measures. This perspective sets the foundation for the subsequent review of related works.

\subsection{SUT-Agnostic Approaches to Fidelity}
\emph{Domain randomization} applies random perturbations (lighting, textures, object placement) during training to encourage models to ignore irrelevant variation; however, it can inadvertently remove task-critical structure that the SUT needs~\cite{tobin2017domain}.
\emph{Game-engine and physically-based rendering (PBR) synthesis} can generate large-scale, diverse scenes but does not guarantee that the SUT will process synthetic and real images through equivalent internal pathways~\cite{richter2016playing,cabonVirtualKITTI22020}.
\emph{Image-translation} methods such as SimGAN and CycleGAN learn mappings between synthetic and real domains to reduce visual discrepancy; yet without explicit semantic constraints, they may alter safety-relevant content (e.g., removing a pedestrian or shifting lane boundaries)~\cite{shrivastava2017simgan,zhu2017cyclegan}. 
\emph{Diffusion models with spatial control} (e.g., ControlNet) condition generation on semantic layouts or edge maps, improving structural fidelity to scene geometry; however, they do not verify whether the SUT's internal reasoning transfers~\cite{rombachHighResolutionImageSynthesis2022,zhangAddingConditionalControl2023}. 
Metric analyses further caution that popular perceptual scores (LPIPS, FID) correlate poorly with downstream task performance~\cite{Zhang2018LPIPS}. 
While valuable, all these methods are SUT-agnostic: they improve visual similarity without checking whether the SUT relies on the same causal evidence motivating SUT-specific approaches.

\subsection{SUT-Specific Fidelity Metrics}
Recognizing the limitations of agnostic metrics, some work has moved towards SUT-specific fidelity. Studies like \cite{haq2021can,codevilla2018offline} measured fidelity by comparing the SUT's performance on real and synthetic data, which essentially corresponds to the OV-fidelity defined in \cite{chengInstanceLevelSafetyAwareFidelity2024}. Beyond OV-fidelity, \cite{chengInstanceLevelSafetyAwareFidelity2024} further introduced a spectrum of fidelity metrics spanning the input, hidden, and output layers of the SUT, assuming a neural network architecture. Moreover, it formalizes instance-level safety-aware fidelity (SA) and shows that calibrating generators to safety outcomes (i.e., those outputs that have actual safety effects on the whole AV system) aligns better with the purpose of AV testing. However, none of these metrics capture \emph{mechanism parity} (whether the SUT made its decision for the same underlying reasons, not just whether it produced the same final output. This necessitates a decisive-feature metric).

As summarized in Table~\ref{tab:fidelity_contrast}, these established metrics have known blind spots.
\emph{IV fidelity} fails when the SUT is locally unrobust: small input perturbations invisible to pixel metrics can cause large output changes~\cite{Szegedy2014Intriguing,chengInstanceLevelSafetyAwareFidelity2024}, so high IV-fidelity does not guarantee high OV-fidelity.
\emph{OV fidelity} can mask mechanism shifts: the SUT may produce identical outputs from different causal evidence (spurious correlation).
\emph{LF fidelity}, which compares broad internal activations, conflates decisive and non-decisive features~\cite{Adebayo2018SanityChecks}; a synthetic image may activate similar neurons overall yet rely on entirely different evidence for the final decision.
These gaps motivate a metric that specifically targets the \emph{decisive} feature subspace, that is, the subset of features that actually influence the SUT's output.

\begin{table}[t]
  \centering
  \caption{Fidelity metrics, what they compare, and typical blind spots.}
  \label{tab:fidelity_contrast}
  \footnotesize
  \setlength{\tabcolsep}{3.5pt}
  \renewcommand{\arraystretch}{1}
  \begin{tabular}{@{}lccc@{}}
    \toprule
    \textbf{Notion} & \textbf{Comparison Target} & \textbf{SUT-Specific?} & \textbf{Blind spot} \\
    \midrule
    IV & pixels/appearance & No  & unrobust $F$ \\
    OV & task outputs      & Yes & spurious evidence \\
    LF & broad internals   & Yes & non-decisive channels \\
    DFF & decisive features& Yes & unrobust XAI (mitigated) \\
    \bottomrule
  \end{tabular}
\end{table}

\section{Foundation}
\label{sec:foundations}
We denote the SUT by a function \( F: \mathbb{R}^{d_0} \to \mathbb{R}^{d_L} \) mapping $d_0$-dimensional inputs to $d_L$-dimensional outputs, and the XAI explainer by \(\Halg\).
We let $x_r$ and $x_s$ denote real and synthetic inputs, respectively.
\(\Dist\) represents the distance metric used to quantify the difference between explanations.
Decisive features are evaluated through the maps $\Halg(F(\cdot))$ and compared using $\Dist$.

\subsection{Scenario Description and Generator}
\label{sec:scenario-generator}
Let $\mathbb{A}$ denote the universe of semantic attributes (e.g., actors and environment factors). A scenario description is $SD=\{(a_k,v_k)\}_{k=1}^K$ with attributes $a_k\in\mathbb{A}$ and values $v_k$ (discrete or continuous, encoded in $\R$ as needed). A synthetic generator $\Gth$ takes $SD$ and produces a sample $x_s\!=\!\Gth(SD)\in\R^{d_0}$; $\Gth$ may be stochastic.
Let \RW{SD} denote the (ideal) set of real samples consistent with $SD$.
In practice one only has a collected dataset $X^{cr}$ and an empirical approximation $\XhR(SD)=X^{cr}\cap \mathrm{RW}(SD)$.

\subsection{Established Fidelity Metrics}
\label{sec:established-fidelity}
We introduce the formal definitions of fidelity metrics from~\cite{chengInstanceLevelSafetyAwareFidelity2024} that are related to our DFF in this subsection.
Let $\Din$ and $\Dout$ compute the distance on $\R^{d_0}$ and $\R^{d_L}$, and let $\eps_{\mathrm{in}},\eps_{\mathrm{out}}>0$.

\paragraph*{Input-Value (IV) Fidelity}
A synthetic $x_s \!=\! \Gth(SD)$ is $\langle \Din, \eps_{\mathrm{in}}\rangle$-IV-fidelitous if
\begin{equation}
\exists\, x_r \in \RW{SD}:\; \Din(x_s, x_r) \le \eps_{\mathrm{in}} .
\label{eq:iv_fidelity}
\end{equation}
Since we only possess the empirical set $\XhR(SD)$, we evaluate Eq.~\eqref{eq:iv_fidelity} by minimizing over $\XhR(SD)$ instead of $\RW{SD}$. Note that this upper-bounds the true minimum (a conservative approximation).

\paragraph*{Output-Value (OV) Fidelity}
Given $F$, $x_s \!=\! \Gth(SD)$ is $\langle F, \Dout, \eps_{\mathrm{out}}\rangle$-OV-fidelitous if
\begin{equation}
\exists\,x_r \in \RW{SD}:\; \Dout\!\bigl(F(x_s), F(x_r)\bigr) \le \eps_{\mathrm{out}}.
\label{eq:ovf_fidelity}
\end{equation}

\paragraph*{Latent-Feature (LF) Fidelity} Beyond input/output matching, one can assess whether the SUT processes synthetic inputs through \emph{similar internal mechanisms}. Given a deep neural network (DNN) $F$ with intermediate layers indexed by $\ell$, and using a distance function $\Dist_{\mathrm{lf}}$ and tolerance $\eps_{\mathrm{lf}}$:
\begin{equation}
\exists\,x_r\in \RW{SD}:\;
\forall\,\ell\in\mathcal{I}_{\ell},\;
\Dist_{\mathrm{lf}}\!\big(F^{(\ell)}(x_s),F^{(\ell)}(x_r)\big)\le \eps_{\mathrm{lf}},
\label{eq:lf_fidelity}
\end{equation}
where $\mathcal{I}_{\ell}$ denotes the set of selected intermediate layer indices of the DNN.

\section{Decisive-Feature Fidelity (DFF)}
\label{sec:method}
Our framework, illustrated in Fig.~\ref{fig:dff-pipeline}, consists of three main components: (1) \textbf{Metric Definition}: We formalize DFF to quantify mechanism parity alongside IV and OV fidelity. (2) \textbf{Estimation}: An XAI explainer extracts decisive features from both real and synthetic inputs to compute the DFF metric. (3) \textbf{Calibration}: A feedback loop uses the DFF metric to optimize the generator's parameters $\Theta$, minimizing the mechanism gap. The following subsections detail each component.

\subsection{DFF Metric Definition}
\label{sec:dff-def}
Given a scenario description $SD$, a fixed SUT $F$, and per-SUT thresholds $(\eps_{\text{in}}, \eps_{\text{out}}, \eps_{\text{dff}})$, we say that a synthetic sample $x_s \sim \Gth(SD)$ is \emph{acceptable for virtual testing} if there exists at least one real $x_r \in \RW{SD}$ such that:
\begin{enumerate}
    \item IV and OV constraints \eqref{eq:iv_fidelity}--\eqref{eq:ovf_fidelity} are satisfied.
    \item The decisive-feature distance satisfies $\Dist \bigl(\Halg(F(x_{s})),\,\Halg(F(x_{r}))\bigr)\le \eps_{\text{dff}}$.
\end{enumerate}
We instantiate decisive features via an XAI explainer $\Halg$ and compare them using a distance function $\Dist$. Note, we exclude LF fidelity ($\varepsilon_{\text{lf}}$) from the top-level acceptance criteria as DFF provides a more specific measure of internal mechanism alignment relevant to the task.
The thresholds $\eps_{\text{in}}$, $\eps_{\text{out}}$, and $\eps_{\text{dff}}$ are user-defined hyperparameters. In practice, they are derived empirically from a calibration set (e.g., using the 95th percentile of distances between matched real pairs) or set based on industry standards (e.g., OV error $<5\%$).
Formally, $x_s$ is $\langle F,\Halg,\Dist,\eps_{\text{dff}}\rangle$-DFF-fidelitous with respect to $x_r$ if
\[
\Dist \bigl(\Halg(F(x_{s})),\,\Halg(F(x_{r}))\bigr)\le \eps_{\text{dff}}.
\]

Over a paired set $\{(x_{r,i},x_{s,i})\}_{i=1}^{N}$, we define the DFF pass-rate as

\begin{equation}
\label{eq:dff_passrate}
\text{Pass-Rate} = \frac{1}{N}\sum_{i=1}^{N} \mathbb{I}\Bigl(
\Dist\bigl(\Halg(F(x_{s,i})),\Halg(F(x_{r,i}))\bigr)\le \eps_{\text{dff}}
\Bigr).
\end{equation}
\[
\mathbb{I}(x)=
\begin{cases}
1, & \text{if } x=\mathrm{TRUE},\\
0, & \text{otherwise}.
\end{cases}
\]

In our experimental evaluation (Table~\ref{tab:dff_thresholds}), we specifically select $\eps_{\text{dff}}$ such that the pass-rate on a reference calibration set is $90\%$ or $95\%$ (denoted as $\eps_{90}$ and $\eps_{95}$).

\paragraph*{Relation to IV/OV/LF}
IV matches pixels, OV matches outputs, LF matches broad internals; DFF matches \emph{decisive} evidence and can diverge under mechanism shifts (right output for wrong reason).

\subsection{DFF Estimator}
\label{sec:def}
We instantiate DFF \emph{exclusively} with a mask-and-infill counterfactual explainer\footnote{Counterfactuals give intervention-style evidence and avoid well-known pitfalls of purely associative saliency, e.g., unrobustness/inconsistency of explanations \cite{ghorbani2019interpretation,huang2023safari,zhao2021baylime}; we therefore choose CF–XAI. That said, conceptually any reliable XAI method can be applied in our DFF framework.} (\emph{CF–XAI}) \cite{FongVedaldi2017Meaningful}.
\textbf{Intuitively CF-XAI works as follows:} Given an image $x$, the explainer asks: ``What is the smallest change to $x$ that would affect the SUT's output?'' It finds a sparse mask $m^\star$ indicating which pixels must be modified (via infill from a learned prior) to change the decision. The resulting \emph{decisive map} $\Halg(F(x))$ highlights these critical regions.

We compute $\Halg(F(x_r))$ and $\Halg(F(x_s))$ independently with the same CF-XAI procedure and define the decisive map by multi-seed averaging:
\begin{equation}
\Halg(F(x)) \coloneqq \frac{1}{K_{\mathrm{cf}}}\sum_{k=1}^{K_{\mathrm{cf}}} m^\star(x;\zeta_k).
\label{eq:avg_map}
\end{equation}
We set $\Dist$ to MSE on pooled $16{\times}16$ averaged maps.

\subsection{DFF-based Calibration Objective}
\label{sec:calib-objective}
We introduce a calibrator network $C_\eta$, parameterized by $\eta$. The goal of the calibrator is to predict optimal generator configuration parameters $\Theta^\ast$ based \emph{only} on the synthetic context, ensuring the method is viable during testing when ground truth is absent.
Given a dataset $D_{\mathrm{paired}}=\{(x_{r,i},x^{(\mathrm{init})}_{s,i},SD_i)\}_{i=1}^{N}$, the calibrator maps the initial synthetic sample and scenario description to the refined parameters: $\Theta_i^\ast=C_\eta(x^{(\mathrm{init})}_{s,i},SD_i)$.
During the \emph{training} of $C_\eta$, we utilize the paired real images $x_{r,i}$ to compute the loss and update $\eta$. We minimize a \emph{combined} loss function that includes input reconstruction loss (IV), SUT output (OV) loss, and our DFF dissimilarity:

\begin{equation}
\label{eq:dff_cal_opt}
\begin{split}
\eta^\ast = \argmin_{\eta} \sum_{i=1}^{N} \Big(
& \Loss_{\text{recon}}(x_{s,i}^\ast, x_{r,i})
+ \beta \,\Loss_{\text{OV}}\!\big(F(x_{s,i}^\ast), F(x_{r,i})\big) \\
& + \lambda_{\text{dff}}\, \Loss_{\text{DFF}}\!\bigl(x_{s,i}^\ast, x_{r,i}\bigr)
\Big),
\end{split}
\end{equation}
where $x_{s,i}^\ast = G_{\Theta_i^\ast}(SD_i)$, and $\Loss_{\text{recon}}$ and $\Loss_{\text{OV}}$ are standard reconstruction (for example $\ell_1$/LPIPS) and task losses (for example MSE). We define $\Loss_{\text{DFF}}(x_s,x_r) \coloneqq \Dist\bigl(\Halg(F(x_s)),\Halg(F(x_r))\bigr)$. We treat $F$ as fixed and non-differentiable and backpropagate gradients only through $C_\eta$ into the generator parameters.

\subsection{Calibration Implementation}
\label{sec:calib-training}

\paragraph*{Calibration knobs and optimizer}
$C_\eta$ predicts generator/post-processing parameters per scenario. We optimize the random seed using Evolution Strategies (ES; population size 32, $\sigma{=}0.1$). For continuous parameters specifically ControlNet scales, guidance scale, and mild post-operations (contrast/brightness/blur) we use Stochastic Gradient Descent (SGD) with learning rates of $5\!\times\!10^{-3}$ and $1\!\times\!10^{-3}$ respectively. Gradients backpropagate through the calibrator and generator parameters only. We treat $F$ as fixed and non-differentiable. The DFF dissimilarity provides the signal for the optimization (SGD or ES) of these generator parameters.

\paragraph*{Algorithm}
We optimize $C_\eta$ to minimize the combined loss in ~\eqref{eq:dff_cal_opt} using SGD/ES on the generator parameters; gradients do not pass through $F$. Decisive maps $\Halg(F(x_r))$ and $\Halg(F(x_s))$ are computed independently for each image in the pair using CF-XAI, then compared via $\Dist$ on pooled representations. Matched pairs or nearest neighbors are used, decisive sets are averaged across counterfactual (CF) seeds, and batching (i.e., computing explanations for $N$ images in parallel) amortizes the computational cost of the CF evaluations.

\begin{algorithm}[t]
\caption{Training the calibrator $C_{\eta}$ (DFF-guided)}
\label{alg:calibration}
\footnotesize
\begin{algorithmic}[1]
\REQUIRE Paired data $\{(x_{r},\,x_{s}^{(\mathrm{init})},\,SD)\}$, generator $G_{\Theta}$, frozen SUT $F$, explainer $\Halg$
\STATE Initialize calibrator $C_{\eta}$
\REPEAT
    \STATE Sample a mini-batch $\mathcal{B}$
    \FOR{each $(x_{r},x_{s}^{(\mathrm{init})},SD)\in\mathcal{B}$}
        \STATE $\Theta^{*} \leftarrow C_{\eta}(x_{s}^{(\mathrm{init})},SD)$
        \STATE $x_{s}^{*} \leftarrow G_{\Theta^{*}}(SD)$
        \STATE $L \leftarrow L_{\mathrm{IV}}(x_{s}^{*},x_{r}) + L_{\mathrm{OV}}(F(x_{s}^{*}),F(x_{r})) + \Dist(\Halg(F(x_{s}^{*})),\Halg(F(x_{r})))$
    \ENDFOR
    \STATE Update $\eta$ to minimize the batch loss (keeping $F$ fixed)
\UNTIL{convergence}
\STATE \textbf{Output:} trained parameters $\eta^{*}$
\end{algorithmic}
\end{algorithm}

\section{Experiments}
\label{sec:experiments}
We evaluate DFF on matched KITTI--VirtualKITTI2 pairs using the three SUTs and address the following two experimental objectives.
\begin{description}[leftmargin=1.1em,labelsep=0.4em,font=\normalfont\bfseries]
  \item[O1 (Assessment):] Does DFF reveal mechanism shifts (high DFF distance) in synthetic samples that otherwise pass standard IV and OV fidelity checks?
  \item[O2 (Calibration):] Does minimizing the DFF objective successfully align decisive features while maintaining non-inferior task performance (OV)?
\end{description}

\subsection{Experimental Setup}
\label{sec:setup}
\textbf{Data and pairing.}
We evaluate on 2{,}126 matched KITTI--VirtualKITTI2 pairs across the five VKITTI2 scenes (Scene01, 02, 06, 18, 20), split 80/20 by scene into 1{,}701 calibration and 425 held-out items.
VirtualKITTI2 provides synthetic ``clone'' sequences that follow the same camera trajectory as the corresponding KITTI raw sequences. We form KITTI--VKITTI2 pairs by matching the scene ID, rendering variant (VirtualKITTI2's ``clone'' condition replicates KITTI lighting and weather), camera (\texttt{Camera\_0}), and frame index.
\textbf{SUTs.} We use three fixed SUT heads: (i) a PilotNet-style steering angle prediction CNN; (ii) YOLOP drivable-area (DA); (iii) YOLOP lane-line (LL). Feature layers are conv5 (steering) and neck pre-decoder (YOLOP), selected for semantic relevance. All SUT weights are frozen (i.e., the SUTs are not trained or fine-tuned; we evaluate their fixed, pre-trained behavior).
\textbf{Metrics.} Details are given in Section~\ref{sec:metrics}.
\textbf{Calibration.} We optimize generator parameters with a combined loss on IV, OV, and DFF, treating the SUT as fixed.
\textbf{Full details.} Optimizers, CF-XAI hyperparameters, and per-layer shapes are documented here: (\repo).
Fidelity thresholds follow domain standards and calibration distributions: $\varepsilon_{\mathrm{out}}$ uses a 0.7 similarity cutoff (e.g., $\mathrm{IoU}\!\ge\!0.7$).

\subsection{Metrics and thresholds}
\label{sec:metrics}

\paragraph*{IV (Input fidelity).}
Recall from Eq.~\eqref{eq:iv_fidelity} that IV fidelity is defined by a distance metric $\Din$. We instantiate $\Din$ using LPIPS \cite{Zhang2018LPIPS} (lower is better).
To facilitate visual comparison in plots where "higher is better," we define a normalized \emph{IV Score} as $1 - \text{LPIPS}_{\text{norm}} \in [0,1]$, where LPIPS is clamped to $[0,1]$. This score is used solely for visualization; the rigorous fidelity check uses the raw LPIPS distance against $\varepsilon_{\text{in}}$.

\paragraph*{OV (Output fidelity / task loss).}
For steering we use an exponential similarity score $\exp(-5 \cdot |\theta_r - \theta_s|) \in [0,1]$, where $\theta$ denotes the predicted angle in radians and the scaling factor 5 normalizes typical angular errors to the $[0,1]$ range (higher is better).
For YOLOP heads we use IoU between predicted masks (higher is better).
We refer to these collectively as OV fidelity; improvements correspond to \emph{increases} in these scores.

\paragraph*{DFF (Decisive-feature distance).}
We compute decisive maps using a mask-and-infill counterfactual explainer applied independently to each image (real and synthetic), then compare the resulting \textbf{pooled decisive maps} by $\mathrm{MSE}$ (lower is better). For each SUT, we choose calibration thresholds $\varepsilon_{90}$ and $\varepsilon_{95}$ as the 90th and 95th percentiles of the DFF distances on the calibration split; these fixed cut-offs are then used for pass/fail decisions on the held-out split and are summarized in Table~\ref{tab:dff_thresholds}.

\begin{table}[t]
  \footnotesize
  \caption{Calibrated DFF thresholds ($\varepsilon$) defined as the 90th and 95th percentiles of DFF distances on the \textbf{calibration split}. Interpretation: $\varepsilon_{90}$ is the threshold below which 90\% of calibration-set pairs fall; a held-out synthetic image passes if its DFF distance is below $\varepsilon$.}
  \label{tab:dff_thresholds}
  \centering
  \footnotesize
  \setlength{\tabcolsep}{6pt}
  \renewcommand{\arraystretch}{1.12}
  \begin{tabular}{@{}lcc@{}}
    \toprule
    {SUT} & {$\varepsilon_{90}$} & {$\varepsilon_{95}$} \\
    \midrule
    Steering   & \epsSteerNinety & \epsSteer \\
    YOLOP--DA  & \epsDANinety    & \epsDA    \\
    YOLOP--LL  & \epsLLNinety    & \epsLL    \\
    \bottomrule
  \end{tabular}
\end{table}
\noindent Full backbone choices, normalization, non-inferiority (NI) margins, and inference details are documented in the repository (\repo).

\subsection{Results and Analysis}
\label{sec:results}

\begin{figure}[!t]
  \centering
  \includegraphics[width=0.7\columnwidth]{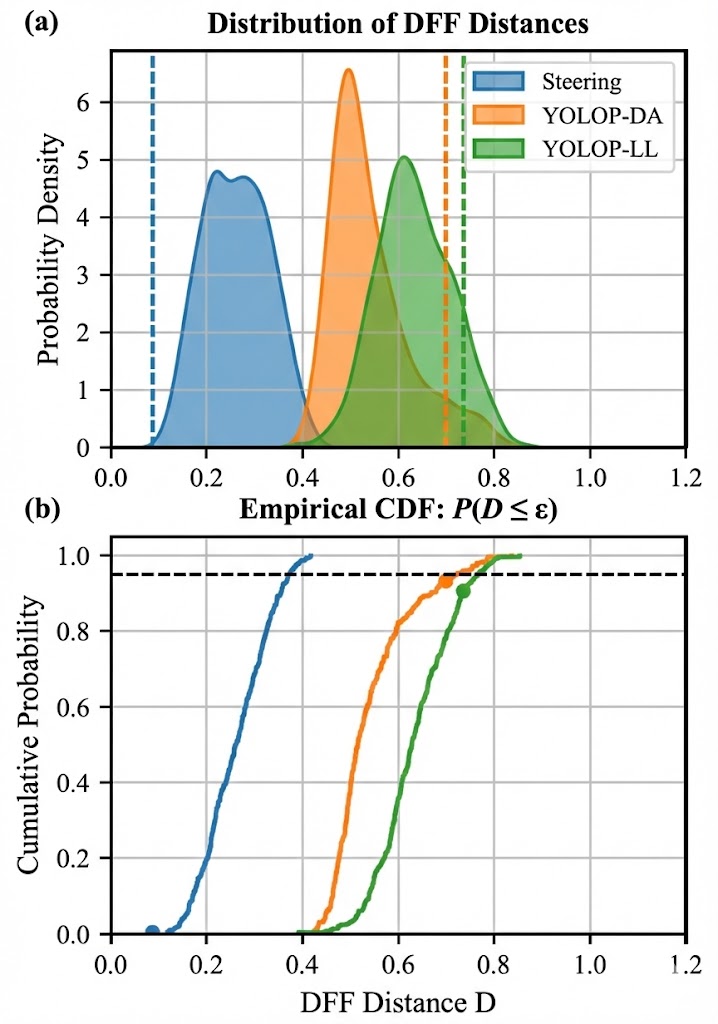}
  \caption{DFF across SUTs. \textbf{(a)} DFF distance $D$ distributions; dashed lines: $\varepsilon_{95}$. \textbf{(b)} CDFs $P(D\le\varepsilon)$; dashed line: 95\%. Curves differ substantially, confirming SUT-specific mechanism mismatch even when OV losses are small (Table~\ref{tab:rq1-pilot}).}
  \label{fig:dff-distance}
\end{figure}

At fixed thresholds determined on the calibration split (Table~\ref{tab:dff_thresholds}), the held-out DFF distributions differ significantly by SUT (Fig.~\ref{fig:dff-distance}). In contrast, OV metrics show weak correlation with these mechanism shifts, as quantified by the low Spearman correlations in Table~\ref{tab:o1_coupling}.

\subsubsection*{O1: Assessment}
\label{sec:o1}
Table~\ref{tab:rq1-pilot} illustrates that OV can improve without aligning decisive features, supporting the need for DFF.
In the table, \emph{Baseline} denotes uncalibrated VirtualKITTI2, \emph{OVF-X} denotes calibration optimizing only OV loss for SUT $X$, and \emph{DFF-X} denotes calibration using the full objective in Eq.~\eqref{eq:dff_cal_opt}.

\begin{table}[t]
  \centering
  \footnotesize
  \caption{Pilot test on one matched pair (Scene20, frame 750). Lower is better for LPIPS and DFF. For OV, we report raw task loss: MSE for steering and $(1{-}\text{IoU})$ for YOLOP heads; lower loss indicates better output agreement. OVF-Steer's high OV loss reflects overfitting to perceptual similarity at the expense of task output.}
  \label{tab:rq1-pilot}
  \footnotesize
  \setlength{\tabcolsep}{5pt}
  \renewcommand{\arraystretch}{1.12}
  \begin{tabular}{@{}l l c c c@{}}
    \toprule
    \textbf{SUT} & \textbf{Variant} & \textbf{IV loss} & \textbf{OV loss}\textsuperscript{$\dagger$} & \textbf{DFF} \\
    \midrule
    Steer & Baseline   & 0.610 & 0.0006  & 0.361 \\
          & OVF-Steer  & 0.473 & 0.3908  & 0.278 \\
          & DFF-Steer  & 0.385 & 0.0028  & 0.229 \\
    \midrule
    DA    & Baseline   & 0.610 & 0.0172 & 0.541 \\
          & OVF-DA     & 0.477 & 0.0044 & 0.630 \\
          & DFF-DA     & 0.390 & 0.0044 & 0.577 \\
    \midrule
    LL    & Baseline   & 0.610 & 0.0019 & 0.712 \\
          & OVF-LL     & 0.421 & 0.0011 & 0.630 \\
          & DFF-LL     & 0.451 & 0.0013 & 0.635 \\
    
    \bottomrule
  \end{tabular}
\end{table}

For \textbf{Steer}, optimizing only OV loss (\emph{OVF-Steer}) improves IV loss (0.610 $\to$ 0.473) and DFF (0.361 $\to$ 0.278), but dramatically worsens OV loss (0.0006 $\to$ 0.3908). This occurs because the OVF optimization can overfit to perceptual similarity at the expense of task-relevant output alignment. The \emph{DFF-Steer} variant achieves the best DFF (0.229) while preserving OV loss near baseline levels (0.0028), demonstrating that DFF-guided calibration successfully aligns decisive features without sacrificing output fidelity.
For \textbf{DA}, OVF improves OV loss (0.0172 $\to$ 0.0044) but worsens DFF (0.541 $\to$ 0.630), illustrating the core problem: output-value parity without mechanism parity. The SUT produces similar outputs but relies on different decisive features. In contrast, \emph{DFF-DA} achieves improved OV loss (0.0172 $\to$ 0.0044) with a smaller DFF increase than OVF-DA (0.541 $\to$ 0.577 vs.\ 0.541 $\to$ 0.630), showing that aligning decisive features can mitigate mechanism divergence while benefiting output fidelity.
For \textbf{LL}, both calibration variants improve over baseline. OVF-LL achieves the best OV loss (0.0019 $\to$ 0.0011) and also improves DFF (0.712 $\to$ 0.630). The \emph{DFF-LL} variant achieves competitive DFF (0.635) with similar OV loss (0.0013). For this SUT, OVF-only calibration yields slightly better DFF than the DFF-guided variant; this pattern persists in the aggregate (Table~\ref{tab:o2_core}), though both variants improve over baseline. This may reflect that for lane-line detection, optimizing output agreement incidentally aligns the decisive features the SUT uses.
These patterns support O1: OV loss can improve while DFF worsens (DA, OVF variant), demonstrating output-value parity without mechanism parity. DFF-guided calibration achieves the lowest DFF for Steering and DA while maintaining competitive OV loss; for LL, OVF-only calibration happens to yield marginally better DFF, though both variants improve substantially over baseline. The DA case most clearly illustrates the failure mode DFF is designed to reveal: OVF-DA reduces OV loss by 74\% yet increases DFF by 16\%, meaning the SUT produces more similar outputs but relies on more divergent decisive features.
The held-out distribution and CDF of DFF distances show clear SUT-specific variation while OV remains tightly clustered (Fig.~\ref{fig:dff-distance}; lower is better for DFF). The weak coupling is quantified by Spearman correlations in Table~\ref{tab:o1_coupling}, confirming DFF carries information not captured by IV/OV.

\begin{table}[t]
\footnotesize
\caption{Weak coupling between OV and DFF on the held-out split (uncalibrated Baseline variant). Spearman $\rho$ with 95\% CI. Lower $|\rho|$ supports O1.}
\label{tab:o1_coupling}
\centering
\footnotesize
\begin{tabular}{l S[table-format=+1.3] S[table-format=+1.3] l}
\toprule
{SUT} & {$\rho$(IV, OV)} & {$\rho$(OV, DFF)} & {95\% CI} \\
\midrule
Steer  & -0.022 & -0.208 & [-0.40, -0.01] \\
DA     & +0.211 & -0.088 & [-0.28, 0.11] \\
LL     & -0.103 & +0.133 & [-0.08, 0.33] \\
\bottomrule
\end{tabular}
\end{table}

Output similarity does not imply mechanism parity: DFF can remain high even when IV and OV distances are low, revealing mechanism shifts that traditional metrics miss.

\begin{figure*}[t]
  \centering
  \includegraphics[width=0.95\textwidth]{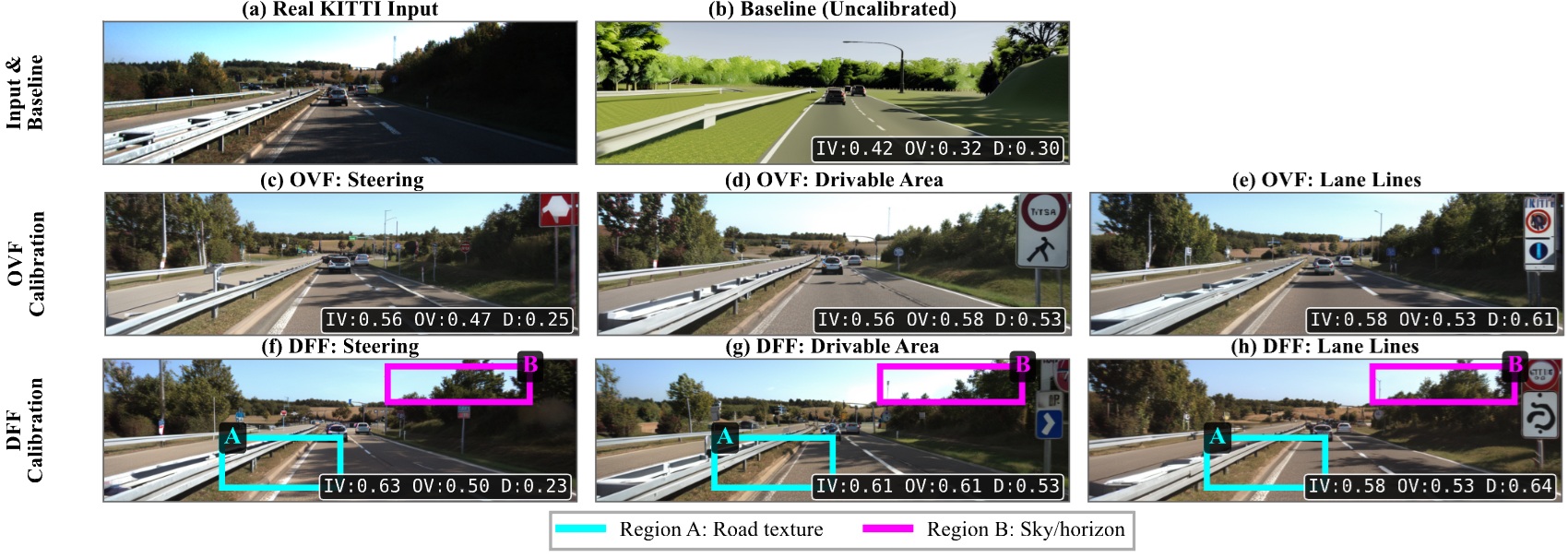}
  \vspace{-10pt}
  \caption{Qualitative comparison of calibration methods on a representative KITTI--Virtual\,KITTI\,2 pair. 
    \textbf{(a)}~Real KITTI input. 
    \textbf{(b)}~Uncalibrated baseline synthetic. 
    \textbf{(c)--(e)}~OVF-calibrated outputs for Steering, Drivable Area, and Lane Lines respectively. 
    \textbf{(f)--(h)}~DFF-calibrated outputs for the same three SUTs. 
    Each synthetic panel displays its IV score (labeled ``IV'', computed as $1-\text{LPIPS}$), OV score, and DFF distance (labeled ``D'').
    Cyan boxes (Region~A) highlight road-surface texture; magenta boxes (Region~B) highlight sky and horizon.}
  \label{fig:qualitative}
\end{figure*}

\begin{figure*}[!t]
  \centering
  \includegraphics[width=0.95\textwidth]{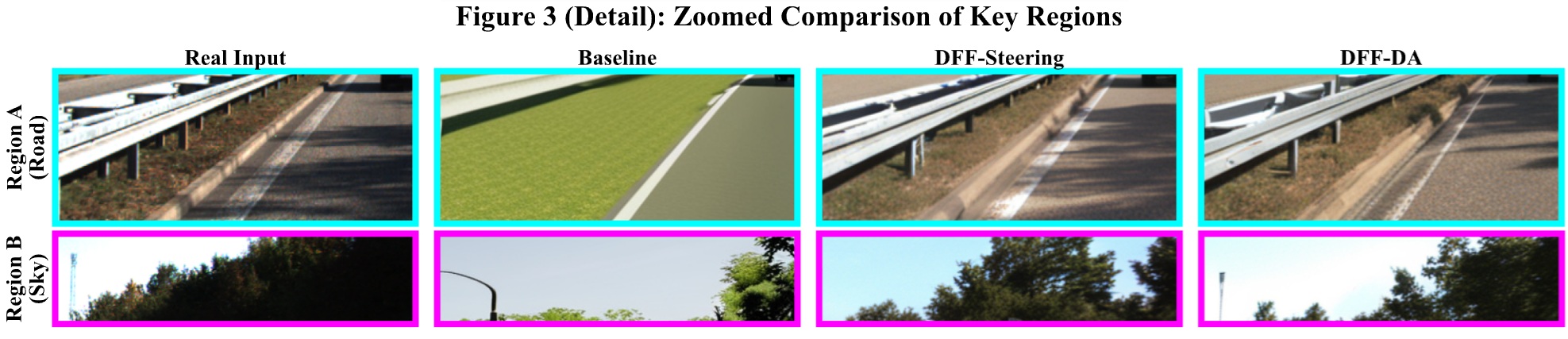}
  \caption{Zoomed comparison of decisive regions from Fig.~\ref{fig:qualitative}. 
    \textbf{Region~A (Road):} Baseline shows unrealistic flat-green rendering; DFF-calibrated variants recover asphalt texture and lane markings. 
    \textbf{Region~B (Sky):} Baseline sky lacks atmospheric variation; DFF calibration restores realistic sky gradients and foliage detail.}
  \label{fig:qualitative-zoom}
\end{figure*}

Figure~\ref{fig:qualitative} illustrates the contrast between optimizing only for outputs and optimizing for decisive features. The OVF variants (second row) reproduce the SUT outputs on the matched scene, yet their visual evidence differs in ways that can shift the SUT’s internal reasoning. DFF calibration produces more realistic road texture and atmospheric rendering in the decisive regions, whereas OVF calibration optimises output similarity without necessarily improving perceptual realism in SUT-critical areas. The DFF-calibrated variants (third row) keep outputs comparable while making the cues that drive the decision, such as road boundaries, lane markings, and vehicle contours, more consistent with the real frame.

The frame in Fig.~\ref{fig:qualitative} contains faint or partially occluded lane markings, which is challenging for YOLOP--LL. We retained this case because it is representative of the dataset's natural difficulty distribution, and it highlights that DFF-guided calibration can improve decisive-feature alignment even when lane cues are weak, a regime where spurious evidence is more likely to mislead the SUT.

\subsubsection*{O2: Calibration}
\label{sec:o2}
Table~\ref{tab:o2_core} summarizes the core effect sizes on the held-out split: DFF-guided calibration reduces DFF (negative $\Delta\text{DFF}$, lower is better) while OV loss is slightly improved on average (positive $\Delta\text{OV}$).
Non-inferiority holds under the pre-declared margins (steering $\delta{=}-0.005$, YOLOP $\delta{=}-0.010$), corresponding to roughly 0.5\% and 1\% relative degradation in OV score.
These thresholds were chosen a priori from pilot experiments using a conservative procedure: we swept calibration settings to induce controlled OV-score drops, identified the smallest drop that produced a consistent and visible downstream degradation in a simple lane-keeping check, and then set $\delta$ strictly below that value.

\begin{table}[t]
\footnotesize
\centering
\caption{O2 core effects on the held-out split. Deltas are computed as (DFF-calibrated $-$ \textbf{Uncalibrated Baseline}).
For IV and OV (similarity, higher is better): positive $\Delta$OV means improvement. 
For DFF (distance, lower is better): negative $\Delta$DFF means improvement.
All changes satisfy pre-declared non-inferiority margins.}
\label{tab:o2_core}
\begin{tabular}{
    l
    S[table-format=+1.3]
    S[table-format=+1.3]
    S[table-format=-1.3]
}
\toprule
\textbf{SUT} & {$\Delta\text{IV}$} & {$\Delta\text{OV}$} & {$\Delta\text{DFF}$} \\
\midrule
Steer & +0.208 & +0.178 & -0.064 \\
DA    & +0.187 & +0.208 & -0.025 \\
LL    & +0.165 & +0.319 & -0.008 \\
\bottomrule
\end{tabular}
\end{table}

 One-sided 95\% CIs for $\Delta\text{OV}$ satisfy the pre-declared margins (steering $\delta=-0.005$, YOLOP heads $\delta=-0.010$); all variants pass NI with $\Delta\text{OV}$ exceeding these thresholds (i.e., $\Delta\text{OV} > \delta$). DFF-calibrated variants do not degrade IV or OV loss and achieve negative $\Delta\text{DFF}$ on the held-out split, indicating improved decisive-feature alignment without sacrificing task performance. The effect sizes are substantial for $\Delta$IV and $\Delta$OV, reflecting meaningful improvements from DFF-guided calibration. The $\Delta$DFF values, while numerically smaller in absolute terms, represent consistent mechanism-alignment gains across all SUTs.

\section{Threats to DFF Validity}
\label{sec:threats}
Our notion of mechanism parity depends on the explanation method, and we currently use a counterfactual-style explainer. All results are on camera-based KITTI and VirtualKITTI2 pairs; multi-sensor and closed-loop settings are left for future work.

\textbf{Qualitative illustration limitations.} Fig.~\ref{fig:qualitative} is provided as an illustrative example to support intuition, but any single qualitative case may be unrepresentative. Therefore, our conclusions are primarily supported by the aggregate quantitative results on the held-out split; the qualitative figure is not used as standalone evidence.

\textbf{Explanation noise.}
CF--XAI is stochastic, which adds noise to DFF and to calibration feedback. We reduce variance via multi-seed averaging (Eq.~\ref{eq:avg_map}), pooling, and sanity checks, and report sensitivity to $K_{\mathrm{cf}}$.

\textbf{Decisive-set identification.}
We use mask-and-infill CF--XAI with sparsity/plausibility constraints, average over CF seeds, and compare $16{\times}16$ pooled maps; stability checks (e.g., model-randomization) help guard against artifacts.

\section{Conclusion}
\label{sec:conclusion}
Virtual testing can fail despite high IV/OV agreement when the SUT achieves similar outputs from different visual evidence. We introduced \emph{Decisive-Feature Fidelity (DFF)} to quantify this \emph{mechanistic realism} by testing whether a fixed SUT relies on consistent decisive features across real and synthetic inputs.
On 2126 KITTI--VirtualKITTI2 pairs, DFF reveals mechanism gaps that IV/OV can miss. On the held-out split, DFF-guided calibration reduces DFF by 0.008--0.064 (MSE on pooled decisive maps) while staying within pre-declared OV non-inferiority margins and improving IV, supporting DFF as an IV/OV complement for \emph{assessment} and, where feasible, \emph{calibration}.
Limitations remain: CF-XAI noise (mitigated via multi-seed averaging and plausibility constraints), conservative pairing assumptions (any hidden trajectory mismatch inflates DFF), and high CF-XAI cost (partly amortized via batching/reuse).
Future work will validate DFF-guided calibration in configurable simulators (CARLA/Prescan) by tuning simulator parameters and OpenDRIVE/OpenSCENARIO specifications, and will develop stronger estimators, extend to broader tasks/datasets (e.g., detection and planning), and provide formal analysis (DFF$\Rightarrow$OV/SA conditions; sample complexity), including sequence-level and distributional extensions.
\noindent\textbf{Closing remark.}
Beyond visual realism, \emph{mechanistic realism} is key for trustworthy virtual testing.

\bibliographystyle{IEEEtran}
\bibliography{refrences/IEEEabrv,refrences/DFF_corrected}

@misc{iso_scenario_2022,
  author  = {{International Organization for Standardization}},
  title   = {{ISO 34502:2022 Road vehicles, Test scenarios for automated driving systems, Scenario based safety evaluation framework}},
  journal = {ISO Standard},
  year    = {2022},
  number  = {ISO 34502:2022}
}

@misc{bsi_scenario_2023,
  author  = {{British Standards Institution}},
  title   = {{BSI Flex 1889 v2.0:2023-09 Formalised natural language description of scenarios for automated driving systems. Specification}},
  journal = {BSI Flex},
  year    = {2023},
  number  = {BSI Flex 1889 v2.0:2023-09}
}

@inproceedings{dosovitskiyCARLAOpenUrban2017,
  author    = {Dosovitskiy, Alexey and Ros, German and Codevilla, Felipe and Lopez, Antonio and Koltun, Vladlen},
  title     = {CARLA: An Open Urban Driving Simulator},
  booktitle = {CoRL},
  year      = {2017},
  volume    = {78},
  pages     = {1--16}
}

@inproceedings{rombachHighResolutionImageSynthesis2022,
  author    = {Rombach, Robin and Blattmann, Andreas and Lorenz, Dominik and Esser, Patrick and Ommer, Bj{\"o}rn},
  title     = {High-Resolution Image Synthesis with Latent Diffusion Models},
  booktitle = {CVPR},
  year      = {2022},
  pages     = {10684--10695}
}

@inproceedings{zhangAddingConditionalControl2023,
  author    = {Zhang, Lvmin and Rao, Anyi and Agrawala, Maneesh},
  title     = {Adding Conditional Control to Text-to-Image Diffusion Models},
  booktitle = {ICCV},
  year      = {2023},
  pages     = {3836--3847}
}

@inproceedings{chengInstanceLevelSafetyAwareFidelity2024,
  author    = {Cheng, Chih-Hong and Wang, Shangguang and Stocco, Daniel and Vingrys, Arminas and Lo Re, Lucia and Lopez, Nuria},
  title     = {Instance-Level Safety-Aware Fidelity Assessment and Adaptive Refinement for Synthetic Driving Data},
  booktitle = {ITSC},
  year      = {2024},
  pages     = {7560--7567}
}

@article{stoccoMindGapStudy2023,
  author  = {Stocco, Andrea and Pulfer, Brian and Tonella, Paolo},
  title   = {Mind the Gap! A Study on the Transferability of Virtual Versus Physical-World Testing of Autonomous Driving Systems},
  journal = {IEEE Transactions on Software Engineering},
  year    = {2023},
  volume  = {49},
  number  = {4},
  pages   = {1928--1940}
}

@inproceedings{johnsonLiteratureReviewSimulation2023,
  author    = {Johnson, Thomas T. and Slattery, John and Bhan, Krishna},
  title     = {Literature Review of Simulation-Based Testing for Autonomous Driving Systems},
  booktitle = {AITest},
  year      = {2020},
  pages     = {23--30}
}

@article{zhao_statistical_2025,
  author  = {Zhao, Xingyu and Aghazadeh-Chakherlou, Robab and Cheng, Chih-Hong and Popov, Peter and Strigini, Lorenzo},
  title   = {On the Need for a Statistical Foundation in Scenario-Based Testing of Autonomous Vehicles},
  journal = {arXiv preprint arXiv:2505.02274},
  year    = {2025}
}

@article{Riedmaier2020Survey,
  author  = {Riedmaier, Stefan and Danquah, Benedict and Ludwig, Johannes and Schick, Boris and Diermeyer, Frank},
  title   = {A Survey on Scenario-Based Safety Assessment of Automated Vehicles},
  journal = {IEEE Trans. on Intelligent Vehicles},
  year    = {2020},
  volume  = {5},
  number  = {2},
  pages   = {233--252}
}

@article{Riedmaier2022Taxonomy,
  author  = {Riedmaier, Stefan and Ludwig, Johannes and Danquah, Benedict and Schick, Boris and Diermeyer, Frank},
  title   = {A Taxonomy and Review of Applications of Scenario-Based Testing of Automated Vehicles},
  journal = {IEEE Transactions on Intelligent Vehicles},
  year    = {2022},
  volume  = {7},
  number  = {4},
  pages   = {732--750}
}

@article{Tang2023ADSTesting,
  author  = {Tang, Peng and Xiong, Yi and Wang, Liang and Liu, Hai},
  title   = {A Survey on Automated Driving System Testing: Landscapes and Trends},
  journal = {ACM Transactions on Software Engineering and Methodology},
  year    = {2023},
  volume  = {32},
  number  = {5},
  pages   = {1--62}
}

@article{Omeiza2022XAI,
  author  = {Omeiza, Daniel and Speakman, Sarah and Cintas, Celia and Weldermariam, Komminist},
  title   = {Smooth Grad-CAM++: An Enhanced Inference Level Visualization Technique for Deep Convolutional Neural Network Models},
  journal = {Information Fusion},
  year    = {2022},
  volume  = {85},
  pages   = {312--327}
}

@inproceedings{tobin2017domain,
  author    = {Tobin, Joshua and Fong, Rachel and Ray, Alex and Schneider, Jonas and Zaremba, Wojciech and Abbeel, Pieter},
  title     = {Domain Randomization for Transferring Deep Neural Networks from Simulation to the Real World},
  booktitle = {IROS},
  year      = {2017},
  pages     = {23--30}
}

@inproceedings{richter2016playing,
  author    = {Richter, Stephan R. and Vineet, Vibhav and Roth, Stefan and Koltun, Vladlen},
  title     = {Playing for Data: Ground Truth from Computer Games},
  booktitle = {ECCV},
  year      = {2016},
  pages     = {102--118}
}

@article{cabonVirtualKITTI22020,
  author  = {Cabon, Yohann and Murray, Naila and Humenberger, Martin},
  title   = {Virtual KITTI 2},
  journal = {arXiv preprint arXiv:2001.10773},
  year    = {2020}
}

@inproceedings{shrivastava2017simgan,
  author    = {Shrivastava, Ashish and Pfister, Tomas and Tuzel, Oncel and Susskind, Josh and Wang, Wenda and Webb, Russell},
  title     = {Learning from Simulated and Unsupervised Images through Adversarial Training},
  booktitle = {CVPR},
  year      = {2017},
  pages     = {2107--2116}
}

@inproceedings{zhu2017cyclegan,
  author    = {Zhu, Jun-Yan and Park, Taesung and Isola, Phillip and Efros, Alexei A.},
  title     = {Unpaired Image-to-Image Translation using Cycle-Consistent Adversarial Networks},
  booktitle = {ICCV},
  year      = {2017},
  pages     = {2223--2232}
}

@inproceedings{Zhang2018LPIPS,
  author    = {Zhang, Richard and Isola, Phillip and Efros, Alexei A. and Shechtman, Eli and Wang, Oliver},
  title     = {The Unreasonable Effectiveness of Deep Features as a Perceptual Metric},
  booktitle = {CVPR},
  year      = {2018},
  pages     = {586--595}
}

@article{haq2021can,
  author  = {Haq, Kazi Zakia and Mathur, Neha and Rojas, Jos{\'e} M.},
  title   = {Can we predict the effectiveness of simulation-based testing? An empirical study on autonomous driving systems},
  journal = {Empirical Software Engineering},
  year    = {2021},
  volume  = {26},
  number  = {5},
  pages   = {90}
}

@inproceedings{codevilla2018offline,
  author    = {Codevilla, Felipe and M{\"u}ller, Matthias and L{\'o}pez, Antonio and Koltun, Vladlen and Dosovitskiy, Alexey},
  title     = {End-to-End Driving via Conditional Imitation Learning},
  booktitle = {ECCV},
  year      = {2018},
  pages     = {501--518}
}

@inproceedings{Szegedy2014Intriguing,
  author    = {Szegedy, Christian and Zaremba, Wojciech and Sutskever, Ilya and Bruna, Joan and Erhan, Dumitru and Goodfellow, Ian and Fergus, Rob},
  title     = {Intriguing Properties of Neural Networks},
  booktitle = {ICLR},
  year      = {2014}
}

@inproceedings{Adebayo2018SanityChecks,
  author    = {Adebayo, Julius and Gilmer, Justin and Muelly, Michael and Goodfellow, Ian and Hardt, Moritz and Kim, Been},
  title     = {Sanity Checks for Saliency Maps},
  booktitle = {NeurIPS},
  year      = {2018},
  pages     = {9525--9536}
}

@inproceedings{ghorbani2019interpretation,
  author    = {Ghorbani, Amirata and Abid, Abubakar and Zou, James Y.},
  title     = {Interpretation of Neural Networks Is Fragile},
  booktitle = {AAAI},
  year      = {2019},
  volume    = {33},
  number    = {1},
  pages     = {3681--3688}
}

@inproceedings{huang2023safari,
  author    = {Huang, Wen and Zhao, Xingyu and Huang, Xiaowei and Robu, Valentin and Flynn, David},
  title     = {SAFARI: Versatile and Efficient Evaluations for Robustness of Interpretability},
  booktitle = {ICCV},
  year      = {2023},
  pages     = {1988--1998}
}

@inproceedings{zhao2021baylime,
  author    = {Zhao, Xingyu and Huang, Wei and Huang, Xiaowei and Robu, Valentin and Flynn, David},
  title     = {BayLIME: Bayesian Local Interpretable Model-Agnostic Explanations},
  booktitle = {UAI},
  year      = {2021},
  volume    = {161},
  pages     = {887--896}
}

@inproceedings{FongVedaldi2017Meaningful,
  author    = {Fong, Ruth C. and Vedaldi, Andrea},
  title     = {Interpretable Explanations of Black Boxes by Meaningful Perturbation},
  booktitle = {ICCV},
  year      = {2017},
  pages     = {3449--3457}
}

@article{10242366,
  author  = {Huang, Zhe and Stocco, Daniel and Mu, Yuting and Tonella, Paolo and Ni, Hongchao and Zhang, Meichen},
  title   = {Testing-Simulation Gap in Autonomous Driving: A Multi-Modal Hierarchical Learning Approach},
  journal = {IEEE Transactions on Intelligent Vehicles},
  year    = {2024},
  volume  = {9},
  number  = {1},
  pages   = {750--760}
}

@article{10588858,
  author  = {Moeini, Saina and Popov, Peter and Huang, Xiaowei and Cheng, Chih-Hong and Strigini, Lorenzo},
  title   = {Safety-Guided Scenario Selection for System-Level Testing of Autonomous Vehicles},
  journal = {IEEE Transactions on Intelligent Transportation Systems},
  year    = {2024},
  volume  = {25},
  number  = {11},
  pages   = {19035--19050}
}

@inproceedings{geigerAreWeReady2012,
  author    = {Geiger, Andreas and Lenz, Philip and Urtasun, Raquel},
  title     = {Are We Ready for Autonomous Driving? The KITTI Vision Benchmark Suite},
  booktitle = {CVPR},
  year      = {2012},
  pages     = {3354--3361}
}

\end{document}